\documentclass{article}

\usepackage[final, nonatbib]{neurips_2021_ml4ps}
\usepackage{settings}
\usepackage{wrapfig}
\usepackage{lipsum}

\title{Arbitrary Marginal Neural Ratio Estimation\\for Simulation-based Inference}

\author{%
    François Rozet \\ 
    University of Liège \\
    \texttt{francois.rozet@uliege.be} \\
    \And
    Gilles Louppe \\
    University of Liège \\
    \texttt{g.louppe@uliege.be} \\
}

\def\knowing{|}

\def\L{\mathcal{L}}
\def\N{\mathcal{N}}

\def\U{\mathcal{U}}
\def\X{\mathcal{X}}

\def\Z{\mathcal{Z}}

\def\E{\mathbb{E}}

\DeclareSIUnit\solarmass{\ensuremath{\mathrm{M}_\odot}}
\DeclareSIUnit\parsec{pc}

\begin{document}

\maketitle

\begin{abstract}
    In many areas of science, complex phenomena are modeled by stochastic parametric simulators, often featuring high-dimensional parameter spaces and intractable likelihoods. In this context, performing Bayesian inference can be challenging. In this work, we present a novel method that enables amortized inference over arbitrary subsets of the parameters, without resorting to numerical integration, which makes interpretation of the posterior more convenient. Our method is efficient and can be implemented with arbitrary neural network architectures. We demonstrate the applicability of the method on parameter inference of binary black hole systems from gravitational waves observations.
\end{abstract}

\section{Introduction}

Formally, a simulator is a stochastic forward model that takes a vector of parameters $\theta \in \Theta$ as input, samples internally a series $z \in \Z \sim p(z \knowing \theta)$ of latent variables and finally produces an observation $x \in \X \sim p(x \knowing \theta, z)$ as output, thereby defining an implicit likelihood $p(x \knowing \theta)$. This likelihood typically is \emph{intractable} as it corresponds to $p(x \knowing \theta) = \int_{\Z} p(x, z \knowing \theta) \d{z}$, the integral of the joint likelihood $p(x, z \knowing \theta)$ over \emph{all} possible trajectories through the latent space $\Z$. In Bayesian inference, we are interested in the posterior
\begin{equation} \label{eq:bayesian inference}
    p(\theta \knowing x) = \frac{p(x \knowing \theta) p(\theta)}{p(x)} = \frac{p(x \knowing \theta) p(\theta)}{\int_{\Theta} p(x \knowing \theta') p(\theta') \d{\theta'}}
\end{equation}
for some observation(s) $x$ and a prior $p(\theta)$, which not only involves the intractable likelihood $p(x \knowing \theta)$ but also an intractable integral over the parameter space $\Theta$. The omnipresence of this problem gave rise to a rapidly expanding field of research \cite{cranmer2020frontier} commonly referred to as \emph{simulation-based} inference. Recent approaches \cite{greenberg2019automatic, hermans2020likelihood, gonccalves2020training, green2021complete} are to learn a surrogate model $\hat{p}(\theta \knowing x)$ of the posterior and, then, proceed as if the latter was tractable.

However, domain scientists are not always interested in the full set of simulator parameters at once. In particular, when interpreting posterior predictions, they generally study several small parameter subsets, like singletons or pairs, while ignoring the others. This task corresponds to estimating the marginal posterior $p(\theta_a \knowing x) = \int_{\Theta_b} p(\theta \knowing x) \d{\theta_b}$ over parameter subspaces $\Theta_a \leq \Theta$ of interest, while the complement subspaces $\Theta_b: \Theta_a \times \Theta_b = \Theta$ are unobserved. To this end, most applications \cite{gonccalves2020training, green2021complete} resort to numerical integration of a surrogate $\hat{p}(\theta \knowing x)$ of the full posterior, which is computationally expensive if $\Theta_b$ is large.

A solution to get rid of numerical integration is to learn directly a surrogate $\hat{p}(\theta_a \knowing x)$ by considering $\theta_b$ as part of the latent variables. If we are interested in a single or a few predetermined subspaces, this approach is reasonable and leads to accurate estimation of marginal posteriors \cite{delaunoy2020lightning, miller2021truncated}. However, if we need to choose \emph{arbitrarily} the subspace $\Theta_a$ at inference time, this solution is not viable anymore as there exists an exponential number ($2^{\dim(\Theta)} - 1$) of marginal posteriors.

\paragraph{Contribution} We build upon neural ratio estimation (NRE) \cite{cranmer2015approximating, hermans2020likelihood} to enable integration-less marginal posterior estimation over arbitrary parameter subspaces. The key idea is to introduce, as input of the ratio estimator, a binary mask $a \in \cbk{0, 1}^{\dim(\Theta)}$ indicating the current subspace $\Theta_a$ of interest. Intuitively, this allows the network to distinguish the subspaces and, thereby, to learn a different ratio for each of them. Our method, dubbed arbitrary marginal neural ratio estimation (AMNRE), can be implemented with arbitrary neural network architectures, including multi-layer perceptrons (MLPs) \cite{hornik1989multilayer} and residual networks \cite{he2016deep}. AMNRE is an amortized method, meaning that inference is simulation-free and can be repeated several times with distinct observations, without retraining. The counterpart is that AMNRE could require a lot of training simulations to produce accurate predictions. The implementation is available at \url{https://github.com/francois-rozet/amnre}.

\paragraph{Related work} Imputation methods \cite{yoon2018gain, belghazi2019learning, li2020acflow} were the first to introduce a binary mask to condition networks with respect to which features are missing. This trick allowed to train a single generative network for all combinations of missing features. Our method differs in that it does not generate likely replacements for the missing features but evaluates the likeliness, conditionally to an observation, of those that are provided.

\section{Arbitrary marginal neural ratio estimation}

\paragraph{NRE} The principle of NRE \cite{hermans2020likelihood} is to train a classifier network $d_\phi: \Theta \times \X \mapsto \sbk{0, 1}$ to discriminate between pairs $(\theta, x)$ equally sampled from the joint distribution $p(\theta, x)$ and the product of the marginals $p(\theta) p(x)$. Formally, the optimization problem is
\begin{equation} \label{eq:nre objective}
    \phi^* = \arg\min_\phi \underset{p(\theta, x)p(\theta')}{\E} \sbk[\big]{ \L(d_\phi(\theta, x)) + \L(1 - d_\phi(\theta'\!, x)) },
\end{equation}
where $\L(p) = - \log p$ is the negative log-likelihood. For this task, the decision function modeling the Bayes optimal classifier \cite{hermans2020likelihood} is
\begin{equation}
    d(\theta, x) = \frac{p(\theta, x)}{p(\theta, x) + p(\theta) p(x)},
\end{equation}
thereby defining the likelihood-to-evidence (LTE) ratio
\begin{equation} \label{eq:lte ratio}
    r(\theta, x) = \frac{d(\theta, x)}{1 - d(\theta, x)} = \frac{p(\theta, x)}{p(\theta) p(x)} = \frac{p(x \knowing \theta)}{p(x)} = \frac{p(\theta \knowing x)}{p(\theta)}.
\end{equation}
Consequently, NRE gives access to an estimator $\log r_\phi(\theta, x) = \logit\rbk*{d_\phi(\theta, x)}$ of the LTE log-ratio and a surrogate $\hat{p}(\theta \knowing x) = r_\phi(\theta, x) p(\theta)$ for the posterior density.

\paragraph{AMNRE} With the additional binary mask $a \in \cbk{0, 1}^{\dim(\Theta)}$, the classifier takes the form $d_\phi(\theta_a, x, a)$ and the optimization problem becomes
\begin{equation} \label{eq:amnre objective}
    \phi^* = \arg\min_\phi \underset{p(\theta, x) p(\theta')}{\E} \, \underset{p(a)}{\E} \sbk[\big]{ \L(d_\phi(\theta_a, x, a)) + \L(1 - d_\phi(\theta_a', x, a)) },
\end{equation}
where $\theta_a = (\theta_i: a_i = 1)$ and $p(a)$ is a mask distribution. In this context, the Bayes optimal classifier (see Appendix \ref{ap:correctness}) is
\begin{equation}
    d(\theta_a, x, a) =\frac{p(\theta_a, x)}{p(\theta_a, x) + p(\theta_a) p(x)},
\end{equation}
meaning that AMNRE gives access to an estimator $\log r_\phi(\theta_a, x, a) = \logit(d_\phi(\theta_a, x, a))$ of all marginal LTE log-ratios and a surrogate $\hat{p}(\theta_a \knowing x) = r_\phi(\theta_a, x, a) p(\theta_a)$ for all marginal posteriors.

AMNRE does not have any particular architectural requirements, with the notable exception of the variable input size of $\theta_a$. To make the method more convenient, $\theta_a$ is replaced by the element-wise product $\theta \cdot a$ ($\theta_a \cdot 1$ and $\theta_b \cdot 0$), carrying the same information at fixed size. The mask $a$ is still required as input since a zero in $\theta \cdot a$ does not unambiguously indicate a zero in $a$. To prevent numerical stability issues when $d_\phi(\theta_a, x, a) \to 1$, the approximate log-ratio $\log r_\phi(\theta_a, x, a)$ is extracted from the neural network and the class prediction is recovered by application of the sigmoid function.

The mask distribution is an important part of AMNRE's training. If some masks $a$ have a small probability $p(a)$ to be selected, it is likely that the estimator will not model their respective marginal posteriors as well as other, more frequent masks. In our experiments, we adopt a uniform mask distribution $p(a) = (2^{\dim(\Theta)} - 1)^{-1}$, leaving the study of this aspect to future work.

\begin{figure}[h]
    \centering
    \includegraphics[scale=0.9]{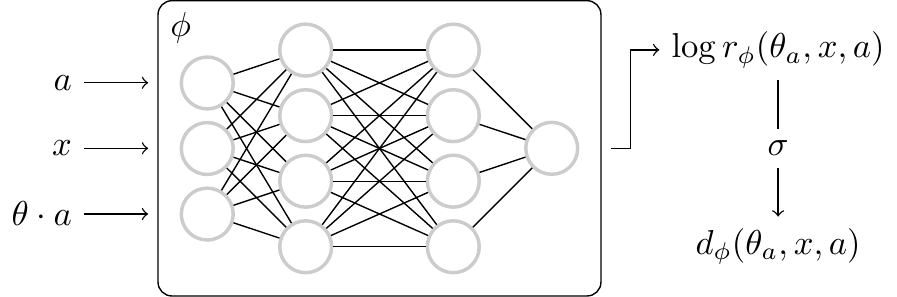}
    \caption{Illustration of AMNRE's classifier architecture.}
    \label{fig:amnre}
\end{figure}

\section{Experiments and results} \label{sec:experiments}

\subsection{Simple likelihood and complex posterior}

\begin{wrapfigure}[27]{r}{0.55\textwidth}
    \centering
    \vspace{-3ex}%
    \includegraphics[scale=0.66]{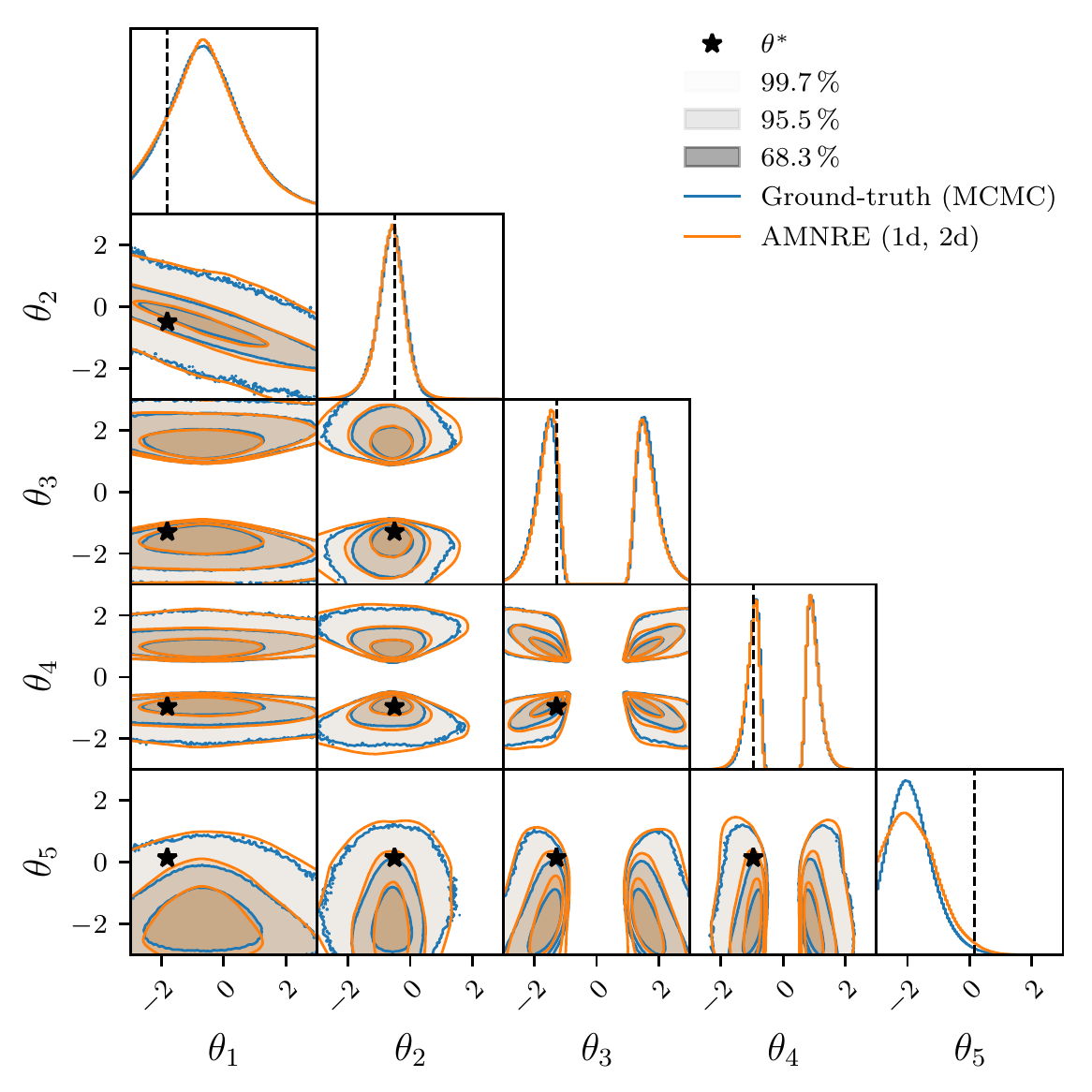}%
    \vspace{-2ex}%
    \caption{Ground-truth posterior against AMNRE 1d and 2d marginal surrogates, for an observation $x^*$ of the SLCP testing set. Density is averaged over three training instances. Contours represent the \SI{68.3}{\percent}, \SI{95.5}{\percent} and \SI{99.7}{\percent} highest posterior density regions. Stars represent the true parameters $\theta^*$ of the observation.}
    \label{fig:slcp gt amnre}
\end{wrapfigure}

\textcite{papamakarios2019sequential} introduce a toy simulator with a 5-dimensional parameter space $\Theta \subseteq \mathbb{R}^5$, for which the likelihood is tractable. Despite its \emph{simple likelihood}, the simulator has a \emph{complex posterior} (SLCP) with four symmetric modes. Hence, SLCP is a non-trivial posterior estimation benchmark that allows to retrieve the ground-truth posterior through Markov chain Monte Carlo (MCMC) sampling \cite{hastings1970monte, chen2012monte} of the likelihood.

We apply AMNRE on SLCP and compare the learned surrogates with the ground-truth posterior. Training details are provided in Appendix \ref{ap:experimental details}. In Figure \ref{fig:slcp gt amnre}, we observe that AMNRE 1d and 2d surrogates are in close agreement with the ground-truth. The structure of the distribution, represented by the credible regions, is modeled correctly, even in low density regions. We also note that the four symmetric modes (see $\theta_3$ and $\theta_4$) are properly recovered, which is sometimes challenging for traditional sampling methods. Concerning the parameter $\theta_5$, we observe that the network is slightly underconfident around the mode, which could indicate that, among the five parameters of SLCP, $\theta_5$ is the hardest to infer. Finally, in Figure \ref{fig:slcp mcmc}, we see that AMNRE is also able to recover the full 5d posterior and the predictions are very consistent with the 1d and 2d surrogates.

\subsection{Gravitational waves}

In recent years, the observations of gravitational waves (GWs) from compact binary coalescences systems have had a massive impact on our understanding of the Universe, partly thanks to inference of the systems' parameters. To obtain posterior samples, the LIGO/Virgo collaboration currently applies MCMC \cite{hastings1970monte, chen2012monte} or nested sampling \cite{skilling2006nested, higson2019dynamic} algorithms to involved physical models of the likelihood of emitted waves \cite{veitch2015parameter, ashton2019bilby}. With these approaches, posterior calculation typically takes days for binary black hole (BBH) mergers and has to be repeated from scratch for each observation.

As a proof of concept, we employ AMNRE to infer the full \num{15}-dimensional set of precessing quasi-circular BBH parameters, given GW observations from the LIGO/Virgo detectors. The simulator details are provided in Appendix \ref{ap:simulators}. After training (see Appendix \ref{ap:experimental details} for details), we evaluate the learned surrogate model on data surrounding GW150914, the first recorded GW event \cite{abbott2016observation}. As reference, we use the posterior samples produced by Bilby \cite{ashton2019bilby} with the \texttt{dynesty} \cite{higson2019dynamic, speagle2020dynesty} nested sampler (MIT License), which leverages the true likelihood. It takes 3 days for Bilby to complete the posterior inference of GW150914, while our network builds histograms (\num{100} bins per dimension) of all 1d and 2d marginal posteriors in about 1 second, on a single 1080Ti GPU.

\begin{figure}[h]
    \centering
    \includegraphics[scale=0.66]{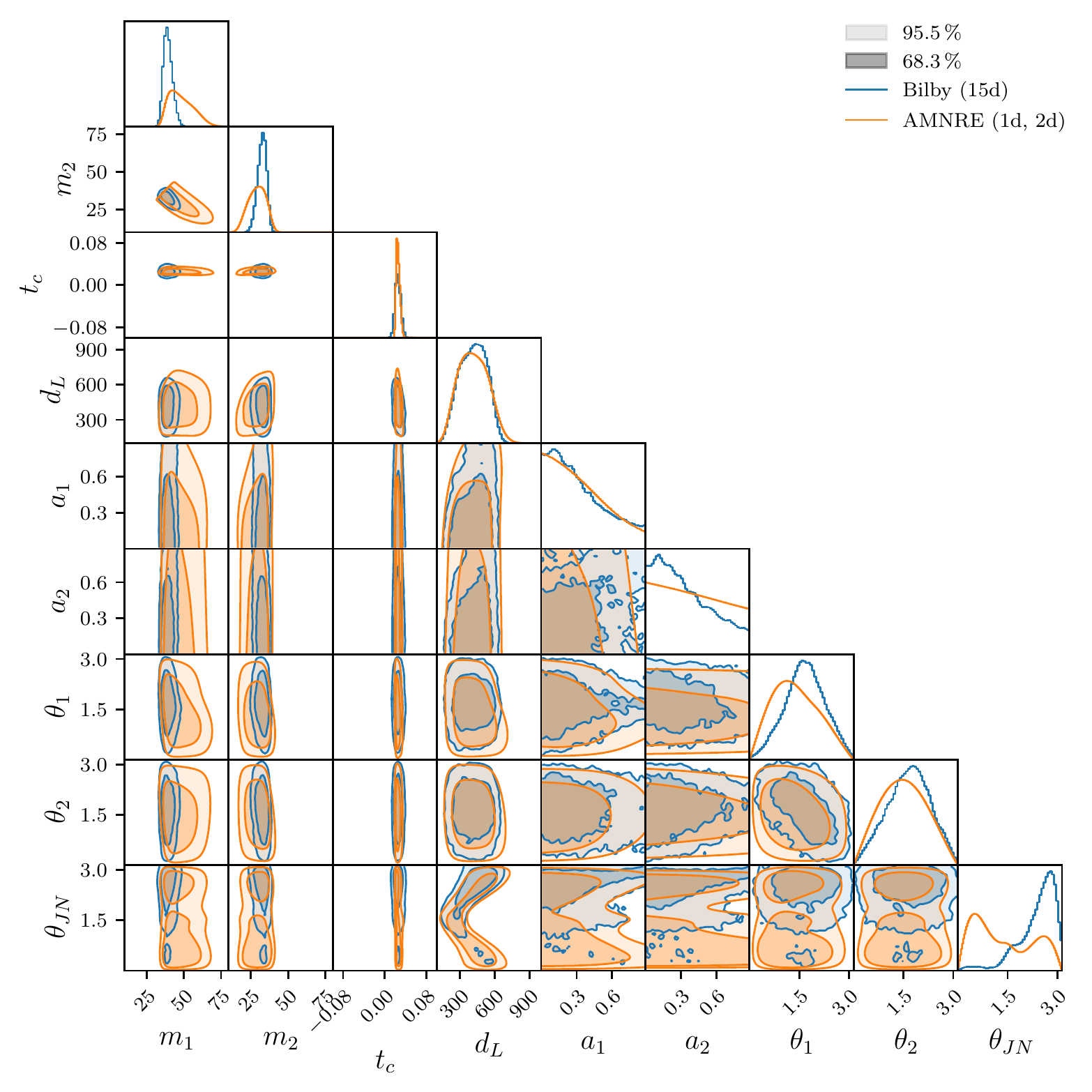}%
    \llap{\raisebox{6.25cm}{%
        \includegraphics[scale=0.66]{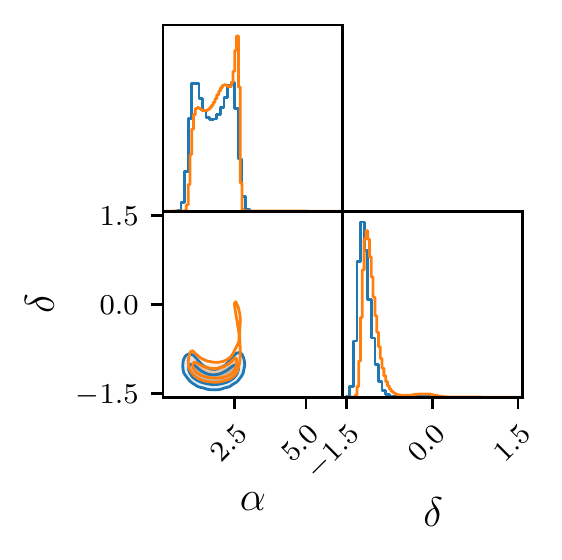}%
        \hspace{1.5cm}
    }}
    \caption{AMNRE 1d and 2d surrogate marginal posteriors against marginalized Bilby posterior samples over a subset of the parameters, for the GW150914 observation. Density is averaged over three training instances.}
    \label{fig:gw bilby amnre}
\end{figure}

As can be seen in Figure \ref{fig:gw bilby amnre}, the surrogate marginal posteriors of AMNRE share the same structure as the marginalized posterior inferred by Bilby. For some parameter subsets, especially those containing the masses $m_1$ and $m_2$ and the inclination angle $\theta_{JN}$, the predictions present significant inaccuracies. For the masses, the surrogates are underconfident but predict the correct modes. For other parameters, including the coalescence time $t_c$, luminosity distance $d_L$ and sky location $(\alpha, \delta)$, the surrogates are in close agreement with Bilby. 

\section{Conclusions} \label{sec:conclusions}

This work introduces AMNRE, a novel simulation-based inference method that enables integration-less marginal posterior estimation over arbitrary parameter subspaces. Through our experiment with the SLCP toy simulator, we demonstrate that the proposed algorithm is indeed able to recover the ground-truth posterior and marginalize it arbitrarily. This experiment also highlights the capacity of AMNRE to model multi-modal distributions, even using a very basic MLP architecture.

The second experiment consists in applying AMNRE to the problem of BBH parameter inference from GW observations. This proof of concept demonstrates that AMNRE is able to analyze GW events several order of magnitude faster (seconds instead of days) than traditional sampling methods. However, if most of the surrogate marginal posteriors seem accurate, some present significant inaccuracies. Possible causes are a lack of estimator expressiveness or insufficient simulation budget; aspects we do not properly study in this work. Still, we believe these results to be a promising demonstration of the applicability of AMNRE for convenient interpretation of the posterior in challenging scientific settings.

\newpage

\begin{ack}
The authors would like to thank Antoine Wehenkel, Arnaud Delaunoy and Joeri Hermans for the insightful discussions and comments. Gilles Louppe is recipient of the ULiège - NRB Chair on Big Data and is thankful for the support of the NRB.
\end{ack}

\printbibliography{}

@article{cranmer2020frontier,
  title={The frontier of simulation-based inference},
  author={Cranmer, Kyle and Brehmer, Johann and Louppe, Gilles},
  journal={Proceedings of the National Academy of Sciences},
  volume={117},
  number={48},
  pages={30055--30062},
  year={2020},
  publisher={National Acad Sciences}
}

@article{cranmer2015approximating,
  title={Approximating likelihood ratios with calibrated discriminative classifiers},
  author={Cranmer, Kyle and Pavez, Juan and Louppe, Gilles},
  journal={arXiv preprint arXiv:1506.02169},
  year={2015}
}

@inproceedings{hermans2020likelihood,
  title={Likelihood-free mcmc with amortized approximate ratio estimators},
  author={Hermans, Joeri and Begy, Volodimir and Louppe, Gilles},
  booktitle={International Conference on Machine Learning},
  pages={4239--4248},
  year={2020},
  organization={PMLR}
}

@article{delaunoy2020lightning,
  title={Lightning-Fast Gravitational Wave Parameter Inference through Neural Amortization},
  author={Delaunoy, Arnaud and Wehenkel, Antoine and Hinderer, Tanja and Nissanke, Samaya and Weniger, Christoph and Williamson, Andrew R and Louppe, Gilles},
  journal={arXiv preprint arXiv:2010.12931},
  year={2020}
}

@inproceedings{greenberg2019automatic,
  title={Automatic posterior transformation for likelihood-free inference},
  author={Greenberg, David and Nonnenmacher, Marcel and Macke, Jakob},
  booktitle={International Conference on Machine Learning},
  pages={2404--2414},
  year={2019},
  organization={PMLR}
}

@article{green2021complete,
  title={Complete parameter inference for GW150914 using deep learning},
  author={Green, Stephen R and Gair, Jonathan},
  journal={Machine Learning: Science and Technology},
  volume={2},
  number={3},
  pages={03LT01},
  year={2021},
  publisher={IOP Publishing}
}

@article{gonccalves2020training,
  title={Training deep neural density estimators to identify mechanistic models of neural dynamics},
  author={Gon{\c{c}}alves, Pedro J and Lueckmann, Jan-Matthis and Deistler, Michael and Nonnenmacher, Marcel and {\"O}cal, Kaan and Bassetto, Giacomo and Chintaluri, Chaitanya and Podlaski, William F and Haddad, Sara A and Vogels, Tim P and others},
  journal={Elife},
  volume={9},
  pages={e56261},
  year={2020},
  publisher={eLife Sciences Publications Limited}
}

@inproceedings{yoon2018gain,
  title={Gain: Missing data imputation using generative adversarial nets},
  author={Yoon, Jinsung and Jordon, James and Schaar, Mihaela},
  booktitle={International Conference on Machine Learning},
  pages={5689--5698},
  year={2018},
  organization={PMLR}
}

@article{belghazi2019learning,
  title={Learning about an exponential amount of conditional distributions},
  author={Belghazi, Mohamed Ishmael and Oquab, Maxime and LeCun, Yann and Lopez-Paz, David},
  journal={arXiv preprint arXiv:1902.08401},
  year={2019}
}

@inproceedings{li2020acflow,
  title={ACFlow: Flow models for arbitrary conditional likelihoods},
  author={Li, Yang and Akbar, Shoaib and Oliva, Junier},
  booktitle={International Conference on Machine Learning},
  pages={5831--5841},
  year={2020},
  organization={PMLR}
}

@article{miller2021truncated,
  title={Truncated Marginal Neural Ratio Estimation},
  author={Miller, Benjamin Kurt and Cole, Alex and Forr{\'e}, Patrick and Louppe, Gilles and Weniger, Christoph},
  journal={arXiv preprint arXiv:2107.01214},
  year={2021}
}

@article{hornik1989multilayer,
  title={Multilayer feedforward networks are universal approximators},
  author={Hornik, Kurt and Stinchcombe, Maxwell and White, Halbert},
  journal={Neural networks},
  volume={2},
  number={5},
  pages={359--366},
  year={1989},
  publisher={Elsevier}
}

@inproceedings{he2016deep,
  title={Deep residual learning for image recognition},
  author={He, Kaiming and Zhang, Xiangyu and Ren, Shaoqing and Sun, Jian},
  booktitle={Proceedings of the IEEE conference on computer vision and pattern recognition},
  pages={770--778},
  year={2016}
}

@inproceedings{ioffe2015batch,
  title={Batch normalization: Accelerating deep network training by reducing internal covariate shift},
  author={Ioffe, Sergey and Szegedy, Christian},
  booktitle={International conference on machine learning},
  pages={448--456},
  year={2015},
  organization={PMLR}
}

@inproceedings{papamakarios2019sequential,
  title={Sequential neural likelihood: Fast likelihood-free inference with autoregressive flows},
  author={Papamakarios, George and Sterratt, David and Murray, Iain},
  booktitle={The 22nd International Conference on Artificial Intelligence and Statistics},
  pages={837--848},
  year={2019},
  organization={PMLR}
}

@article{hastings1970monte,
  title={Monte Carlo sampling methods using Markov chains and their applications},
  author={Hastings, W Keith},
  year={1970},
  publisher={Oxford University Press}
}

@book{chen2012monte,
  title={Monte Carlo methods in Bayesian computation},
  author={Chen, Ming-Hui and Shao, Qi-Man and Ibrahim, Joseph G},
  year={2012},
  publisher={Springer Science \& Business Media}
}

@article{skilling2006nested,
  title={Nested sampling for general Bayesian computation},
  author={Skilling, John},
  journal={Bayesian analysis},
  volume={1},
  number={4},
  pages={833--859},
  year={2006},
  publisher={International Society for Bayesian Analysis}
}

@article{higson2019dynamic,
  title={Dynamic nested sampling: an improved algorithm for parameter estimation and evidence calculation},
  author={Higson, Edward and Handley, Will and Hobson, Michael and Lasenby, Anthony},
  journal={Statistics and Computing},
  volume={29},
  number={5},
  pages={891--913},
  year={2019},
  publisher={Springer}
}

@article{speagle2020dynesty,
  title={dynesty: a dynamic nested sampling package for estimating Bayesian posteriors and evidences},
  author={Speagle, Joshua S},
  journal={Monthly Notices of the Royal Astronomical Society},
  volume={493},
  number={3},
  pages={3132--3158},
  year={2020},
  publisher={Oxford University Press}
}

@article{veitch2015parameter,
  title={Parameter estimation for compact binaries with ground-based gravitational-wave observations using the LALInference software library},
  author={Veitch, John and Raymond, Vivien and Farr, Benjamin and Farr, Will and Graff, Philip and Vitale, Salvatore and Aylott, Ben and Blackburn, Kent and Christensen, Nelson and Coughlin, Michael and others},
  journal={Physical Review D},
  volume={91},
  number={4},
  pages={042003},
  year={2015},
  publisher={APS}
}

@article{ashton2019bilby,
  title={BILBY: a user-friendly Bayesian inference library for gravitational-wave astronomy},
  author={Ashton, Gregory and H{\"u}bner, Moritz and Lasky, Paul D and Talbot, Colm and Ackley, Kendall and Biscoveanu, Sylvia and Chu, Qi and Divakarla, Atul and Easter, Paul J and Goncharov, Boris and others},
  journal={The Astrophysical Journal Supplement Series},
  volume={241},
  number={2},
  pages={27},
  year={2019},
  publisher={IOP Publishing}
}

@article{khan2016frequency,
  title={Frequency-domain gravitational waves from nonprecessing black-hole binaries. II. A phenomenological model for the advanced detector era},
  author={Khan, Sebastian and Husa, Sascha and Hannam, Mark and Ohme, Frank and P{\"u}rrer, Michael and Forteza, Xisco Jim{\'e}nez and Boh{\'e}, Alejandro},
  journal={Physical Review D},
  volume={93},
  number={4},
  pages={044007},
  year={2016},
  publisher={APS}
}

@article{bohe2016phenompv2,
  title={PhenomPv2--technical notes for the LAL implementation},
  author={Boh{\'e}, Alejandro and Hannam, Mark and Husa, Sascha and Ohme, Frank and P{\"u}rrer, Michael and Schmidt, Patricia},
  journal={LIGO Technical Document, LIGO-T1500602-v4},
  year={2016}
}

@article{abbott2016observation,
  title={Observation of gravitational waves from a binary black hole merger},
  author={Abbott, Benjamin P and Abbott, Richard and Abbott, TD and Abernathy, MR and Acernese, Fausto and Ackley, Kendall and Adams, Carl and Adams, Thomas and Addesso, Paolo and Adhikari, RX and others},
  journal={Physical review letters},
  volume={116},
  number={6},
  pages={061102},
  year={2016},
  publisher={APS}
}

@article{clevert2015fast,
  title={Fast and accurate deep network learning by exponential linear units (elus)},
  author={Clevert, Djork-Arn{\'e} and Unterthiner, Thomas and Hochreiter, Sepp},
  journal={arXiv preprint arXiv:1511.07289},
  year={2015}
}

@article{kingma2014adam,
  title={Adam: A method for stochastic optimization},
  author={Kingma, Diederik P and Ba, Jimmy},
  journal={arXiv preprint arXiv:1412.6980},
  year={2014}
}

@article{loshchilov2017decoupled,
  title={Decoupled weight decay regularization},
  author={Loshchilov, Ilya and Hutter, Frank},
  journal={arXiv preprint arXiv:1711.05101},
  year={2017}
}

@article{loshchilov2016sgdr,
  title={Sgdr: Stochastic gradient descent with warm restarts},
  author={Loshchilov, Ilya and Hutter, Frank},
  journal={arXiv preprint arXiv:1608.03983},
  year={2016}
}

\appendix

\section{Correctness of AMNRE} \label{ap:correctness}

Reformulating \eqref{eq:amnre objective}, we have
\begin{align*}
    L & = \iiint_{\Theta \times \X \times \Theta} p(\theta, x) p(\theta') \underset{p(a)}{\E} \sbk[\big]{\L(d_\phi(\theta_a, x, a) + \L(1 - d_\phi(\theta_a', x, a))} \d{\theta} \d{x} \d{\theta'} \\
    & = \iint_{\Theta \times \X} \underset{p(a)}{\E} \sbk[\big]{p(\theta, x) \L(d_\phi(\theta_a, x, a)) + p(\theta) p(x) \L(1 - d_\phi(\theta_a, x, a))} \d{\theta} \d{x} \\
    & = \underset{p(a)}{\E} \iint_{\Theta_a \times \X} \underbrace{\sbk[\big]{p(\theta_a, x) \L(d_\phi(\theta_a, x, a)) + p(\theta_a) p(x) \L(1 - d_\phi(\theta_a, x, a))}}_{\ell(d_\phi(\theta_a, x, a))} \d{\theta_a} \d{x},
\end{align*}
which is minimized only if each term $\ell(d_\phi(\theta_a, x, a))$ is itself minimized. Assuming $p(\theta_a) p(x) > 0$, if $p(\theta_a, x) = 0$, $\ell$ is uniquely minimized by the value $0$. Otherwise, if $p(\theta_a, x) > 0$, the minimum is only reached by a value $q$ such that
\begin{alignat*}{2}
    && 0 & = \frac{\d{\ell(q)}}{\d{q}} \\
    &&& = p(\theta_a, x) \frac{\d{\L(q)}}{\d{q}} + p(\theta_a) p(x) \frac{\d{\L(1 - q)}}{\d{q}} \\
    &&& = p(\theta_a, x) \frac{-1}{q} + p(\theta_a) p(x) \frac{1}{1 - q} \\
    \Leftrightarrow \quad && q & = \frac{p(\theta_a, x)}{p(\theta_a, x) + p(\theta_a) p(x)} = d(\theta_a, x, a) .
\end{alignat*}
Importantly, if $p(\theta_a, x) = 0$, $d(\theta_a, x, a) = 0$ and still minimizes $\ell$. Therefore, as long as $p(\theta_a) p(x) > 0$, $d(\theta_a, x, a)$ is the optimal classifier.

\section{Simulators} \label{ap:simulators}

\subsection{Simple likelihood and complex posterior}

In this toy simulator, $\theta \in \mathbb{R}^5$ parametrizes a 2d multivariate Gaussian from which four points are independently sampled to construct an observation $x$. The generative process \cite{papamakarios2019sequential} is
\begin{align*}
    \theta_i & \sim \U(-3, 3) \quad \text{for} \quad i = 1, \dots, 5 \\
    s_1 & = \theta_3^2, \quad s_2 = \theta_4^2, \quad \rho = \tanh(\theta_5) \\
    \mu & = (\theta_1, \theta_2), \quad \Sigma = \begin{pmatrix} s_1^2 & \rho s_1 s_2 \\ \rho s_1 s_2 & s_2^2 \end{pmatrix} \\
    x & = (z_1, \dots, z_4) \quad \text{where} \quad z_j \sim \N(\mu, \Sigma),
\end{align*}
for which the likelihood $p(x \knowing \theta) = \prod_j p(z_j \knowing \theta)$ is tractable.

\subsection{Gravitational waves}

As we do not have enough knowledge in the domain, we borrow the BBH simulator implemented by \textcite{green2021complete} (MIT License). We succinctly describe the generative process in this section. For more information, please refer to the original paper or the implementation.

\paragraph{Prior} We perform inference over the full \num{15}-dimensional set of precessing quasi-circular BBH parameters: component masses $(m_1, m_2)$, reference phase $\phi_c$, coalescence time $t_c$, luminosity distance $d_L$, spin magnitudes $(a_1, a_2)$, spin angles $(\theta_1, \theta_2, \phi_{12}, \phi_{JL})$, inclination angle $\theta_{JN}$, polarization angle $\psi$, and sky location $(\alpha, \delta)$. To analyze GW150914, we take a prior uniform over
\begin{alignat*}{2}
    \SI{10}{\solarmass} & \leq m_i && \leq \SI{80}{\solarmass} \\
    \SI{-0.1}{\second} & \leq t_c && \leq \SI{0.1}{\second} \\
    \SI{100}{\mega\parsec} & \leq d_L && \leq \SI{1000}{\mega\parsec} \\
    \num{0} & \leq a_i && \leq \num{0.88}
\end{alignat*}
and standard over the remaining quantities. We take $t_c = 0$ to be the trigger time of GW150914 and constraint $m_1 \geq m_2$.

\paragraph{Waveform generation} The simulator generates waveforms using the \texttt{IMRphenomPv2} frequency-domain processing model \cite{khan2016frequency, bohe2016phenompv2} and assumes stationary Gaussian noise with respect to the noise power spectral density (PSD) estimated from \SI{1024}{s} of detector data prior to GW150914. The frequency ranges from \num{20} to \SI{1024}{\hertz} and each waveform has a duration of \SI{8}{\second}. The waveforms are whitened with respect to the estimated PSD.

\paragraph{Waveform processing} The observations are quite large (\num{16384} features) and, thereby, impractical to store on disk and feed to a neural network. To alleviate this problem, the waveforms are compressed to a reduced-order basis corresponding to the first \num{128} components of a singular value decomposition (SVD). Using more SVD components did not help producing better predictions, likely due to the higher ratio of noise in less significant components.

Since an observation corresponds to two waveforms from two geographically distant detectors (H1 and L1) and frequency-domain signals are represented by complex-valued vectors, each processed observation is a vector of $128 \times 2 \times 2 = 512$ real-valued numbers.

\paragraph{Noise} For the training set, the noise of the detectors is not added to the stored waveforms. Instead, noise realizations are sampled with respect to the PSD in real time during training, which effectively increases the size of the training set.

\section{Experimental details} \label{ap:experimental details}

\paragraph{Datasets} For each simulator, we use three fixed datasets of pairs $(\theta, x) \sim p(\theta, x)$ to train, validate and test AMNRE, respectively. The sizes of the datasets are provided in Table \ref{tab:datasets}.

\begin{table}[h]
    \centering
    \caption{Dataset sizes for each simulator.}
    \begin{tabular}{lccc}
        \toprule
        Simulator & Training set & Validation set & Testing set \\
        \midrule
        SLCP & \num{1048576} & \num{131072} & \num{131072} \\
        GW & \num{4194304} & \num{131072} & \num{131072} \\
        \bottomrule
    \end{tabular}
    \label{tab:datasets}
\end{table}

\paragraph{Architectures} For SLCP, we use an MLP with 7 hidden layers of \num{256} neurons and ELU \cite{clevert2015fast} activation functions. For GW, the classifier is a residual residual network \cite{he2016deep} consisting of \num{17} residual blocks of 2 linear layers with 512 neurons and ELU \cite{clevert2015fast} activation functions. In the blocks, we insert batch normalization layers \cite{ioffe2015batch} before the activation functions.

\paragraph{Training} All networks are optimized with the AdamW \cite{kingma2014adam, loshchilov2017decoupled} stochastic optimization algorithm. At each epoch, the batches are built by sampling without replacement from the training set. The independent parameters $\theta'$ are obtained by shifting circularly ($i \gets i + 1$ and $n \gets 1$) the batch of parameters $\theta$. Each element in the batch has a different mask, sampled from the uniform mask distribution.

For SLCP, we apply a \enquote{reduce on plateau} scheduling to the learning rate, that is, we divide the learning rate by a factor \num{2} each time the loss on the validation set has not decreased for \num{7} consecutive epochs. The training stops when the learning rate reaches \num{e-6} or lower. For GW, we apply a learning rate cosine annealing \cite{loshchilov2016sgdr} over \num{512} epochs. Other hyperparameters are provided in Table \ref{tab:hyperparameters}.

\begin{table}[H]
    \centering
    \caption{Training hyperparameters.}
    \begin{tabular}{lcc}
        \toprule
        Hyperparameter & SCLP & GW \\
        \midrule
        Optimizer & AdamW & AdamW \\
        Weight decay & \num{e-4} & \num{e-4} \\
        Batch size & \num{1024} & \num{1024} \\
        Batches per epoch & \num{256} & \num{1024} \\
        Epochs & - & \num{512} \\
        Scheduling & reduce on plateau & cosine annealing \\
        Initial learning rate & \num{e-3} & \num{2e-4} \\
        Final learning rate & \num{e-6} & \num{e-6} \\
        \bottomrule
    \end{tabular}
    \label{tab:hyperparameters}
\end{table}

\section{Additional figures}

\begin{figure}[h]
    \centering
    \includegraphics[scale=0.66]{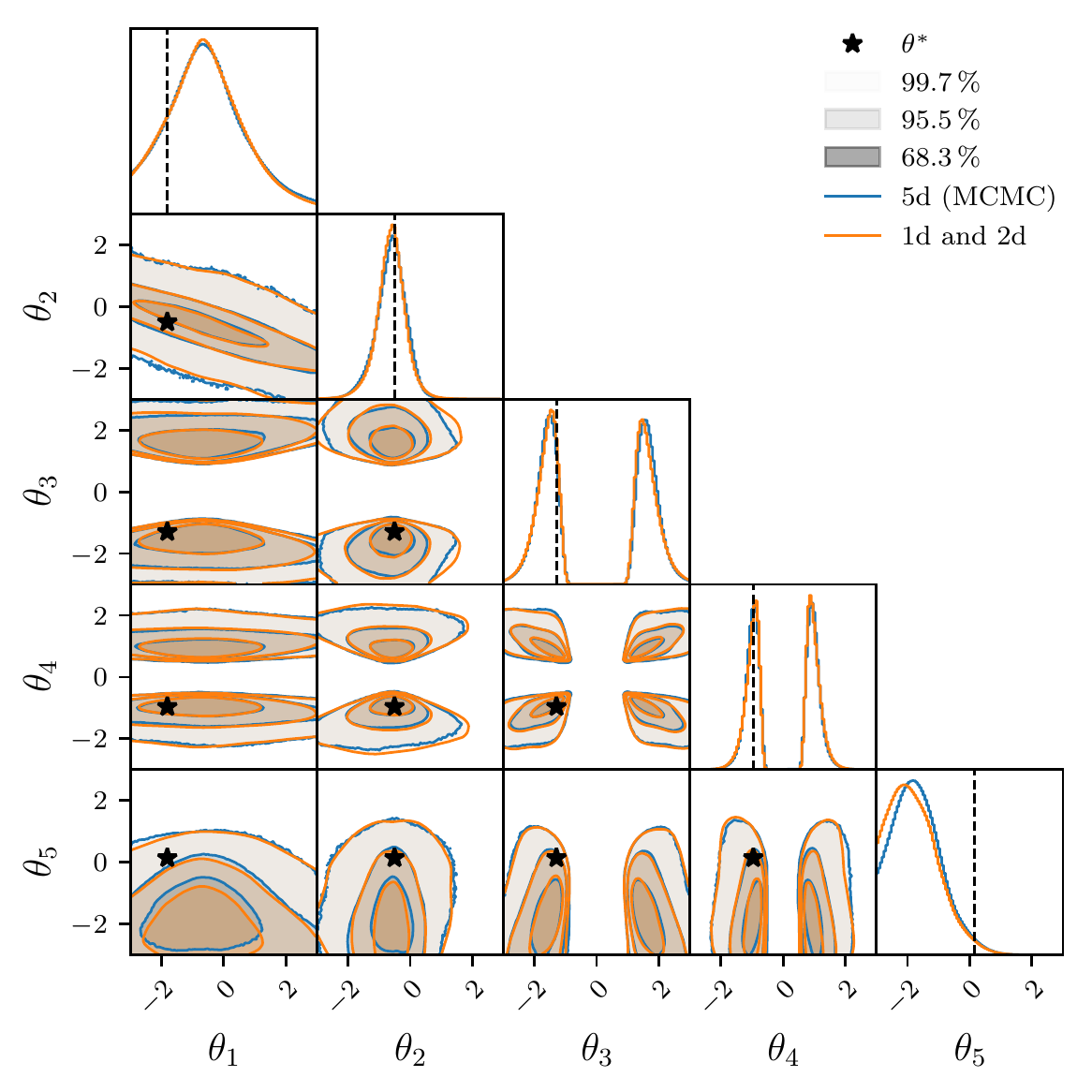}
    \caption{AMNRE full 5d surrogate posterior against 1d and 2d marginal surrogates, for an observation of the SLCP testing set. The predictions of AMNRE for the marginal posteriors are consistent with the predictions for the full posterior, marginalized onto the 1d and 2d subspaces.}
    \label{fig:slcp mcmc}
\end{figure}

\begin{figure}[h]
    \begin{center}
        \includegraphics[width=0.495\textwidth]{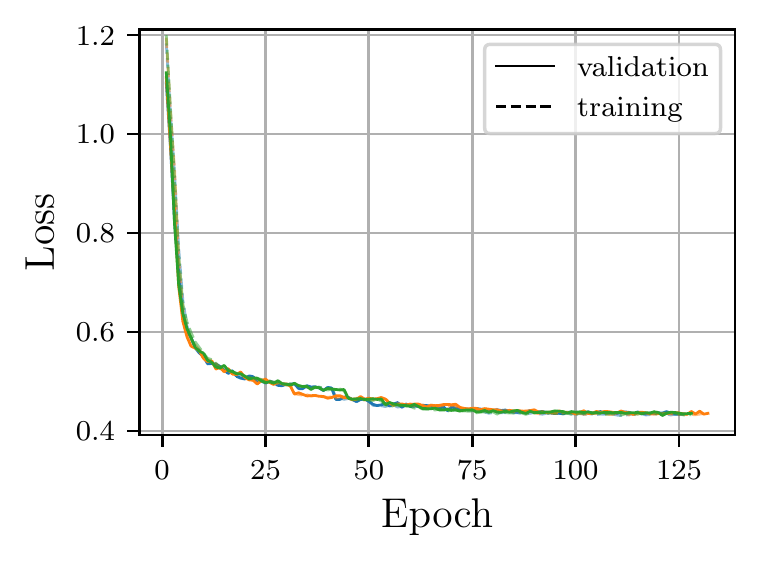}%
        \includegraphics[width=0.495\textwidth]{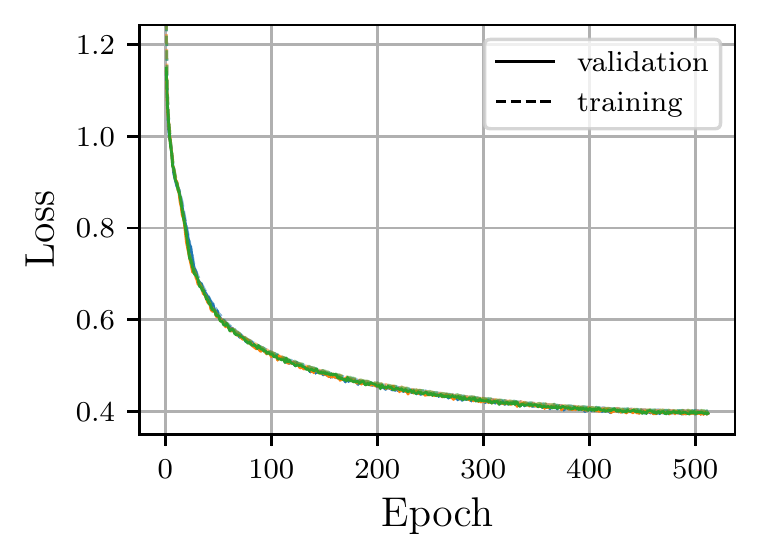}%
        \vspace{-2ex}
    \end{center}
    \caption{Mean training and validation losses of AMNRE surrogate models for SLCP (left) and GW (right) simulators. Each color corresponds to a different training instance. All instances converge without signs of overfitting. Training takes around 5 minutes for SLCP and 8 hours for GW, on a single 1080Ti GPU.}
    \label{fig:losses}
\end{figure}

\begin{figure}[h]
    \centering
    \includegraphics[width=0.495\textwidth]{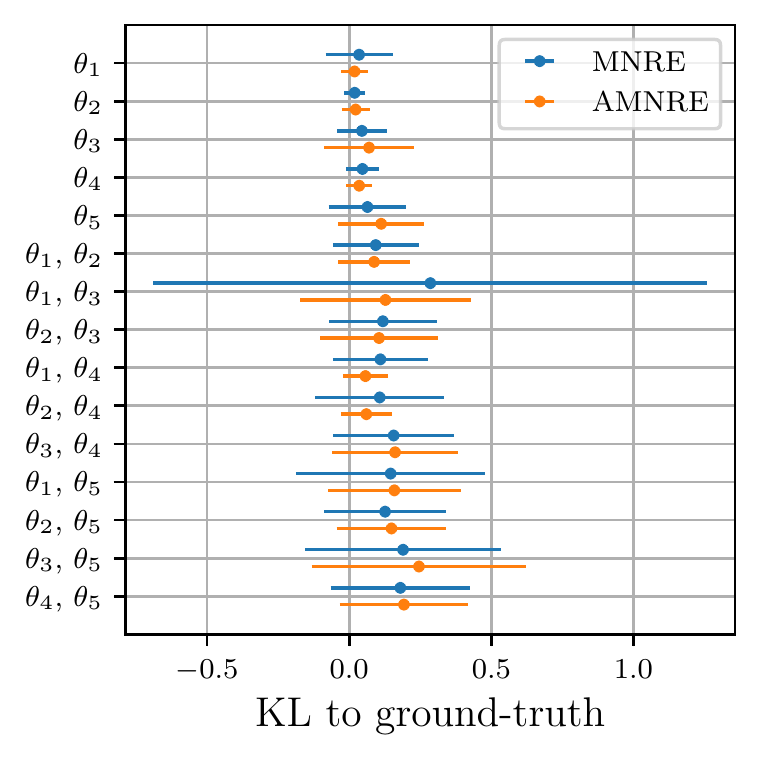}
    \caption{KL divergence to the marginalized ground-truth posterior of 1d and 2d surrogate marginal posterior histograms. The bars represent the mean and standard deviation over 64 observations from the SLCP testing set. AMNRE does not diverges more from the ground-truth than MNRE \cite{hermans2020likelihood, miller2021truncated}, despite using a single network for all subspaces.}
    \label{fig:slcp divergence}
\end{figure}

\begin{figure}[h]
    \begin{center}
        \includegraphics[width=0.495\textwidth]{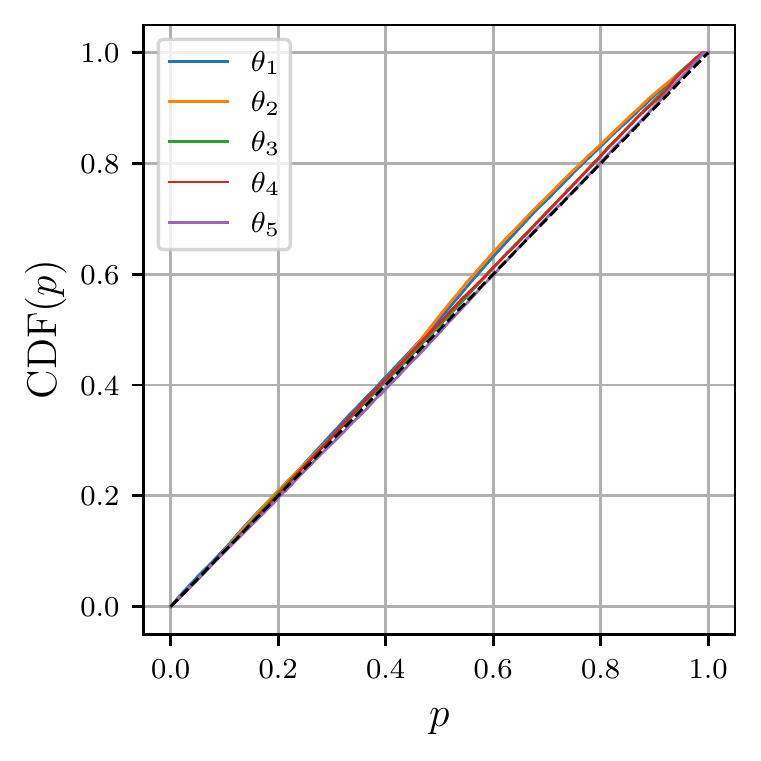}%
        \includegraphics[width=0.495\textwidth]{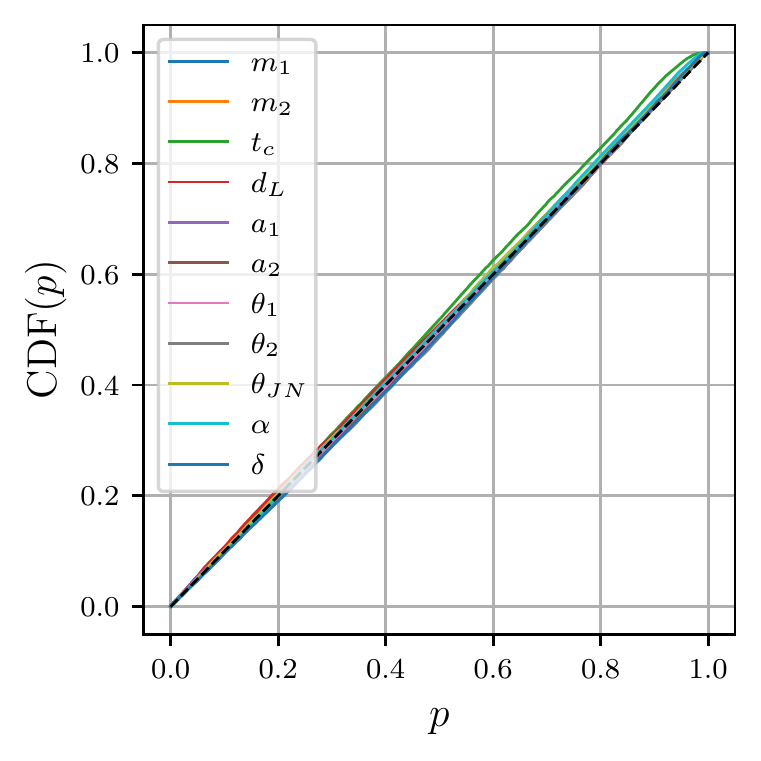}%
        \vspace{-2ex}
    \end{center}
    \caption{Cumulative distribution function (CDF) of the percentiles $p$ of \num{8192} parameters $\theta^*$ in the one-dimensional surrogate marginal posteriors $\hat{p}(\theta_i \knowing x^*)$ for pairs $(\theta^*, x^*)$ of the SLCP (left) and GW (right) testing sets. If the surrogate posterior is consistent with the prior, \ie{} if $\E_{p(x)} \sbk[\big]{\hat{p}(\theta \knowing x)} \approx p(\theta)$, the percentiles should be distributed uniformly between 0 and 1. Since the CDFs lie close to the diagonal, we conclude that the surrogates are consistent with the prior. \\[1ex]
    Importantly, one \emph{cannot} conclude that the network models properly the posterior from this result, as \emph{any} distribution consistent with the prior, including the prior itself, would present diagonal CDFs. \textcite{green2021complete} inadvertently draw this erroneous conclusion.}
    \label{fig:calibration}
\end{figure}

\end{document}